\documentclass[letterpaper, 10 pt, conference]{ieeeconf}  % Comment this line out if you need a4paper

\IEEEoverridecommandlockouts                              % This command is only needed if 
\usepackage{graphics} % for pdf, bitmapped graphics files
\usepackage{times} % assumes new font selection scheme installed
\usepackage{amsmath} % assumes amsmath package installed
\usepackage{amssymb}  % assumes amsmath package installed
\usepackage{graphicx}
\usepackage{color}
\usepackage{cite}
\usepackage{censor}
\makeatletter
\let\NAT@parse\undefined
\makeatother
\usepackage[colorlinks,linkcolor=blue, citecolor=blue, anchorcolor=blue]{hyperref}
\usepackage{multirow}
\usepackage{booktabs}
\usepackage[nameinlink,noabbrev]{cleveref}
\crefname{figure}{Fig.}{Figs.}
\crefname{equation}{Eq.}{Eqs.}
\crefname{section}{Sec.}{Secs.}
\crefname{table}{Table}{Tables}

\title{\LARGE \bf
MuxGel: Simultaneous Dual-Modal Visuo-Tactile Sensing via Spatially Multiplexing and Deep Reconstruction}

\author{
Zhixian Hu, Zhengtong Xu, Sheeraz Athar, Juan Wachs, Yu She
\thanks{This material is partially based upon work supported by the National Science Foundation under Award 2322056, 2423068, and 2520136, and in part by the U.S. Department of Agriculture under Award 2023-67021-39072 and 2024-67021-42878. Any opinions, findings, and conclusions or recommendations expressed in this material are those of the authors and do not necessarily reflect the views of the funding agencies.}
\thanks{Edwardson School of Industrial Engineering, Purdue University, West Lafayette, IN, USA. \tt\small{\{jpwachs,shey\}@purdue.edu}}
}

\begin{document}

\maketitle
\thispagestyle{empty}
\pagestyle{empty}

%%%%%%%%%%%%%%%%%%%%%%%%%%%%%%%%%%%%%%%%%%%%%%%%%%%%%%%%%%%%%%%%%%%%%%%%%%%%%%%%
\begin{abstract}
High-fidelity visuo-tactile sensing is important for precise robotic manipulation, yet most vision-based tactile sensors rely on opaque coatings that enable tactile sensing but block direct visual observation. We propose MuxGel, a spatially multiplexed sensor that captures both external visual information and contact-induced tactile signals through a single camera. By using a checkerboard coating pattern, MuxGel interleaves tactile-sensitive regions with transparent windows for external vision. This design maintains standard form factors, allowing for plug-and-play integration into GelSight-style sensors by simply replacing the gel pad. To recover dense visual and tactile signals from the multiplexed inputs, we develop a U-Net-based reconstruction framework trained with a sim-to-real pipeline. Experiments on unseen objects demonstrate the framework's generalization and accuracy. We further demonstrate MuxGel in grasping tasks, where visual feedback supports alignment and tactile feedback supports contact interaction. Results show that MuxGel enables single-camera dual-modal sensing within a GelSight-style implementation, providing local visual feedback and reconstructed tactile feedback with potential extension to other optical tactile sensors. Project webpage: \url{https://zhixianhu.github.io/muxgel/}.

\end{abstract}

\section{Introduction}

Effective robotic manipulation requires both vision and touch \cite{cui2021toward,luu2025manifeel,xu2025unit}. Vision provides global context for planning and approach phases, whereas tactile sensing provides local feedback for contact detection\cite{prince2025tacscope}, force regulation\cite{lin2024self}, and fine manipulation\cite{zhou2025hand}. Traditionally, visual feedback is obtained through free-space cameras that observe the manipulation region directly \cite{gao2023hand, zhang2025canonical}. However, as the end-effector approaches an object, the manipulator or object can occlude the contact interface, forcing the system to shift from visual guidance to local tactile feedback during interaction.

\begin{figure}[thpb]
  \centering
  \includegraphics[width=0.46\textwidth]{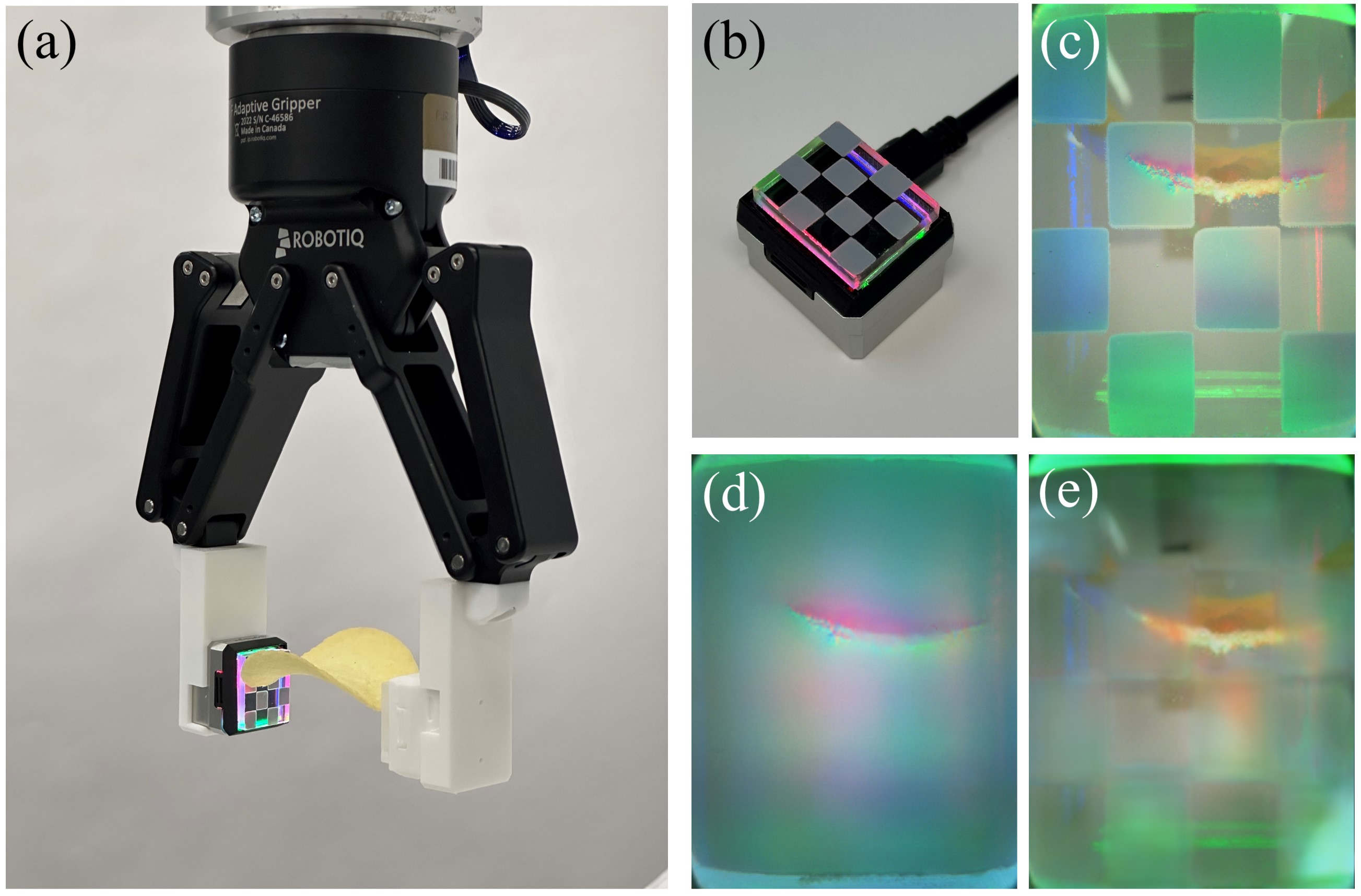}
  \caption{Grasping a chip with MuxGel: (a) A Robotiq gripper grasping a chip using MuxGel for simultaneous visuo-tactile sensing via spatial multiplexing. (b) MuxGel with the 4$\times$4 checkerboard configuration integrated into a GelSight Mini by replacing only the gel pad, without optical or mechanical redesign. (c) Raw multiplexed sensor output. (d) Reconstructed tactile image. (e) Reconstructed visual image.}
  \label{fig:intro}
\end{figure} 

To capture detailed local information, tactile sensors are increasingly embedded into robotic fingertips \cite{hu2023machine}. Vision-based tactile sensors are widely used due to their high spatial resolution, compact mechanical design, and simple wiring. Representative systems include GelSight, which reconstructs surface geometry using photometric cues \cite{yuan2017gelsight}, TacTip, which tracks internal pins \cite{ward2018tactip}, DelTact, which estimates deformation from dense-pattern optical flow \cite{zhang2022deltact}, and DTact, which images contact through semi-transparent and absorptive layers \cite{lin2023dtact}. However, these sensors mainly provide tactile perception. Their internal coatings or markers enable deformation sensing but limit direct observation of the external environment.

Several dual-modal approaches have been proposed to reduce the occlusion gap, while introducing new trade-offs. Adding a dedicated camera alongside a tactile sensor increases the fingertip size and introduces parallax, complicating cross-modal alignment \cite{chaudhury2022using, shimonomura2016robotic}. Alternatively, utilizing transparent elastomers with embedded visual markers reduces bulk \cite{zhang2025design, fan2024vitactip, wang2022spectac, yamaguchi2019tactile}, but sparse markers limit tactile spatial resolution and local contact detail. Mode-switching designs offer another route, but mechanical switches can increase system complexity \cite{dong2026look}. Material-driven sensors reduce this hardware burden by changing transparency through special coatings or gels \cite{athar2023vistac,luu2023soft,nguyen2025vi2tap,hogan2022finger}. However, they still alternate between visual and tactile states and therefore do not provide simultaneous visual and tactile measurements during contact, when contact formation, object motion, and local deformation may need to be monitored together.

\begin{figure}[thpb]
  \centering
  \includegraphics[width=0.46\textwidth]{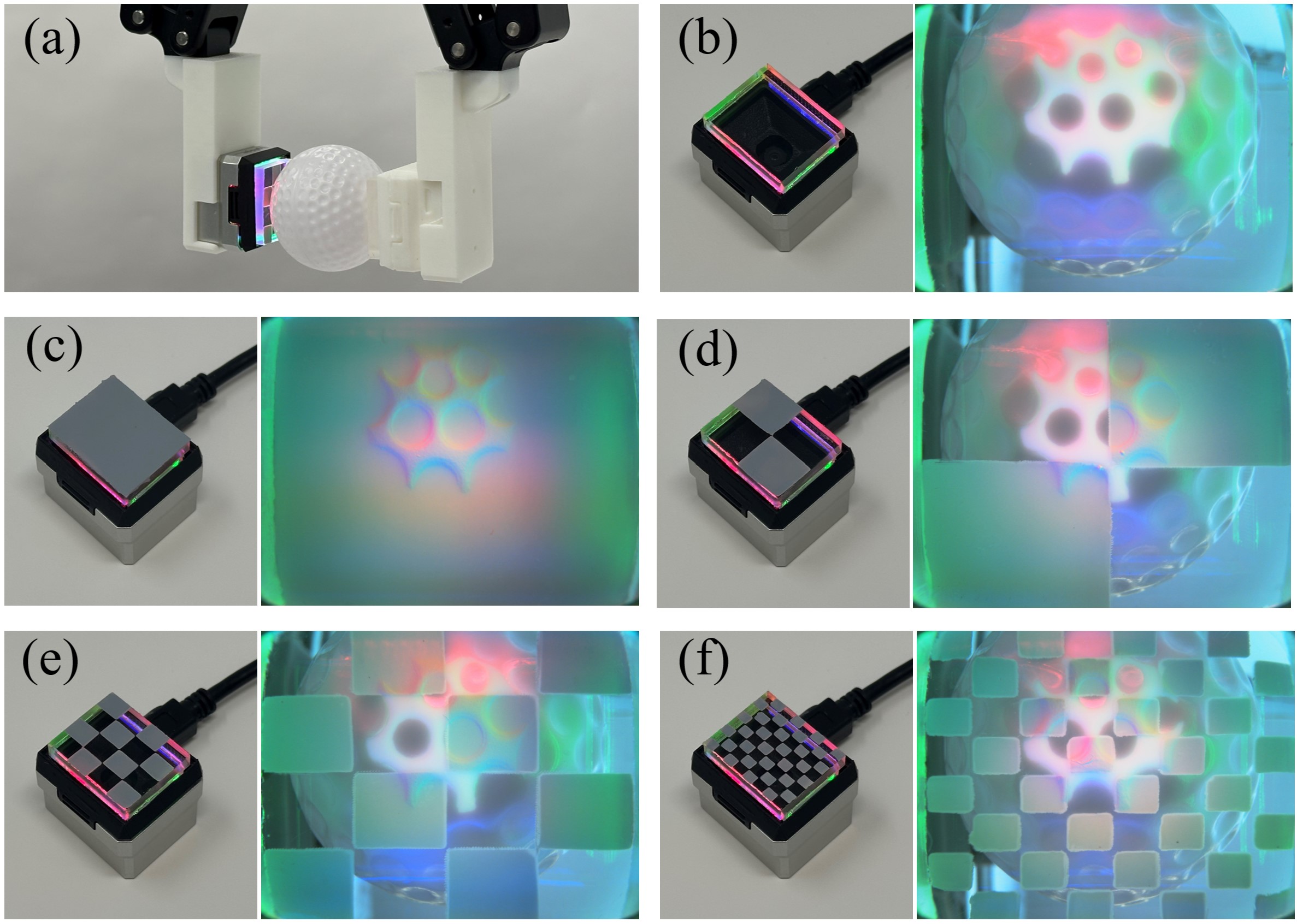}
  \caption{MuxGel configurations and raw observations. (a) A 4$\times$4 sensor integrated with a Robotiq gripper during a grasp. (b)-(f) Hardware patches (left) and captured images (right) for pure vision, pure tactile, and multiplexed (2$\times$2, 4$\times$4, 8$\times$8) configurations.}
  \label{fig:config}
\end{figure} 

To address these limitations, we introduce MuxGel, an integrated hardware and software framework for simultaneous visuo-tactile perception via spatial multiplexing, as shown in \cref{fig:intro}. MuxGel uses a checkerboard coating pattern that assigns different regions to visual appearance sensing and contact deformation. This design preserves the pad geometry and mechanical interface of GelSight-style sensors, enabling integration by replacing only the gel pad. To reconstruct dense dual-modal signals from the multiplexed input, we develop a deep reconstruction pipeline based on a shared ResNet encoder \cite{he2016deep} and two U-Net-style decoders \cite{ronneberger2015u}, drawing on computational vision and image inpainting methods \cite{isola2017image,liu2018image,zhang2018single}. The model first learns from large-scale physics-based simulation and is then fine-tuned with real-world data. Experiments on unseen-object dual-modal reconstruction and robotic grasping demonstrate accurate reconstruction, cross-object generalization, and real-time visuo-tactile feedback for manipulation.

\section{Method \label{sec:method}}
\subsection{Hardware Design}

The MuxGel sensor's capabilities stem from a spatially multiplexed coating strategy.  We follow the standard GelSight gel pad fabrication process, with a modified coating step. Specifically, instead of applying a uniform coating layer, we utilize a checkerboard-pattern mold to selectively apply gray Lambertian paint as a diffuse reflective layer, yielding alternating coated and transparent regions. We fabricate three checkerboard configurations ($2 \times 2$, $4 \times 4$, and $8 \times 8$) to study coating resolution, along with two baseline pads for modality-specific ground truth: a fully coated tactile-only pad and a transparent vision-only pad, as shown in \cref{fig:config}. This modular design allows for rapid pad change without modifications to the underlying sensor hardware.

\subsection{Simulation Data Generation Pipeline}

\begin{figure*}[thpb]
  \centering
  \includegraphics[width=0.92\textwidth]{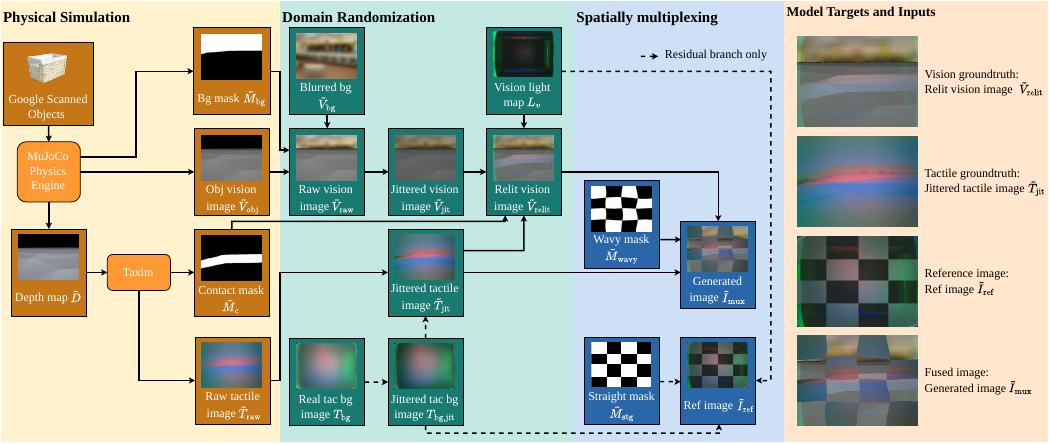}
  \caption{Large-scale physics-based simulation pipeline for visual-tactile data generation. Bg: Background; Obj: Object; Tac: Tactile; Ref: Reference.}
  \label{simData}
\end{figure*} 

To reduce real-world data collection cost, we develop a randomized physics-based simulation pipeline. The pipeline synthesizes realistic spatially multiplexed images ($\tilde{I}_{\text{mux}}$) by simulating the physical deformation, optical properties, and MuxGel masking, as shown in \cref{simData}. The tilde notation ($\tilde{\cdot}$) denotes synthetic data generated by our simulation pipeline, distinguishing it from real sensor data.

We build the simulation environment in MuJoCo \cite{todorov2012mujoco} using models from the Google Scanned Objects dataset \cite{downs2022google} and their MuJoCo conversions \cite{zakka2022scannedobjectsmujoco}. A virtual camera is placed 14 mm from the object surface to match the GelSight Mini geometry, and 50 viewpoints are sampled per object to render RGB images ($\tilde{V}_{\text{obj}}$) and depth maps ($\tilde{D}$).

Tactile interaction is simulated by combining the rendered depth map with Taxim \cite{si2022taxim}, a GelSight Mini simulator with low pixel-wise intensity errors. The depth map $\tilde{D}$ is converted into an object height map and compared with the calibrated elastomer spatial profile to approximate soft deformation. Given a randomly sampled pressing depth in $[0.01, 1.5]$ mm, the intersection produces a contact mask ($\tilde{M}_c$) and a relative contact height field. Following Taxim, surface gradients are mapped to illumination intensities via a calibrated polynomial table. To improve realism, a shadow generation algorithm employs ray-tracing approximations to cast shadows based on the contact geometry, outputting a raw tactile image ($\tilde{T}_\text{raw}$).

To narrow the sim-to-real gap, we apply domain randomization that targets the sensor's optics and shared illumination structure. Since external scenes are often out of focus in vision-based tactile sensors, backgrounds ($\tilde{V}_\text{bg}$) from the IndoorCVPR dataset \cite{quattoni2009recognizing} are blurred with a disk defocus kernel and fused with $\tilde{V}_\text{obj}$ using a background mask ($\tilde{M}_\text{bg}$) derived from the depth and image intensity. The raw vision image $\tilde{V}_\text{raw}$ is formulated as:
\begin{equation}
\tilde{V}_\text{raw} = \tilde{M}_\text{bg} \odot \tilde{V}_\text{bg} + (1 - \tilde{M}_\text{bg}) \odot \tilde{V}_\text{obj},
\label{VisSimRaw}
\end{equation}

To evaluate the decoupling network's ability to handle complex real-world sensor noises, our pipeline supports two tactile target formulations: absolute and residual. In the absolute formulation, the raw simulated tactile image $\tilde{T}_\text{raw}$ is utilized directly. In the residual formulation, we isolate the deformation-induced difference image $\tilde{T}_\text{diff}$ by subtracting the no-contact tactile background
\begin{equation}
\tilde{T}_\text{diff} = \tilde{T}_\text{raw} - \tilde{T}_\text{org,bg},\label{TacDiff}\end{equation}where $\tilde{T}_\text{org,bg}$ is the simulated non-contact tactile background. 

Because external ambient light leakage affects both modalities simultaneously, we apply correlated color jittering. In the absolute formulation, brightness, contrast, saturation, and hue are perturbed jointly for $\tilde{T}_\text{raw}$ and $\tilde{V}_\text{raw}$ to obtain $\tilde{T}_\text{jit}$ and $\tilde{V}_\text{jit}$. In the residual formulation, this joint perturbation is applied to a randomly sampled real tactile background $T_\text{bg}$ and $\tilde{V}_\text{raw}$, yielding $T_\text{bg,jit}$ and $\tilde{V}_\text{jit}$, respectively. The final tactile image $\tilde{T}_\text{jit}$ used for residual learning is then constructed as
\begin{equation}
\tilde{T}_\text{jit} = \tilde{T}_\text{diff} +T_\text{bg,jit}.
\label{t_jitt}
\end{equation}
This correlated jittering mechanism encourages the network to learn fundamental structural differences rather than relying on trivial color shifts.

Moreover, the visual and tactile modalities in MuxGel share an internal optical environment. To simulate this, we utilize real blank sensor images captured in a dark environment as vision light maps ($L_v$), and the processed tactile images ($\tilde{T}_\text{jit}$) as the contact-induced illumination field. We relight the jittered vision image by applying these priors, guided by the contact mask $\tilde{M}_c$:
\begin{equation}
\tilde{V}_\text{relit} = \tilde{M}_c \odot \left(\tilde{V}_\text{jit} \odot \tilde{T}_\text{jit} \right) + (1 - \tilde{M}_c) \odot \left( \tilde{V}_\text{jit}  \odot L_v \right),
\label{VisSimRelit}
\end{equation}
where $\tilde{V}_\text{relit}$ denotes the relit vision image.

The final sensor observation is synthesized with a checkerboard mask. In the real sensor, the checkerboard boundary distortion can arise from elastomer deformation, coating non-uniformity, and lens distortion, and may therefore be contact-dependent. Explicitly modeling this deformation-correlated mask distortion would require additional calibration of the gel mechanics and optical path. As a lower-cost approximation, we generate a random wavy checkerboard mask ($\tilde{M}_\text{wavy}$) with base grids ranging from 2$\times$2 to 8$\times$8 to expose the network to plausible boundary variations caused by manufacturing tolerances, elastomer warping, and lens distortions. The boundaries of the wavy mask $B(t)$ are perturbed by:
\begin{equation}
B(t) = P_\text{base} + A \cdot \sin(2\pi f t + \phi),
\label{SimWavyMask}    
\end{equation}
where $P_\text{base}$ is the nominal boundary position, $A \in [0, 5.0]$ is a randomized amplitude, $f$ is the spatial frequency, and $\phi$ is a phase shift. The spatially multiplexed input $\tilde{I}_\text{mux}$ to the network is then 
\begin{equation}
\tilde{I}_\text{mux} = \tilde{M}_\text{wavy} \odot \tilde{T}_\text{jit} + (1 - \tilde{M}_\text{wavy}) \odot \tilde{V}_\text{relit}.
\label{muxSim}
\end{equation}

To facilitate the decoupling process, particularly for the residual learning formulation, our network requires a spatial reference image ($\tilde{I}_\text{ref}$). This reference image is constructed using a nominal straight checkerboard mask ($\tilde{M}_\text{stg}$) corresponding to the base grid dimensions. The reference image $\tilde{I}_\text{ref}$ acts as an illumination prior and an ideal multiplexing layout by compositing $T_\text{bg,jit}$ and $L_v$:
\begin{equation}\tilde{I}_\text{ref} = \tilde{M}_\text{stg} \odot T_\text{bg,jit} + (1 - \tilde{M}_\text{stg}) \odot L_v.\label{RefImage}\end{equation}

Finally, we require the contact-area ratio in $\tilde{M}_c$ to exceed 5\%, a heuristic threshold that removes near-non-contact samples with little deformation supervision while retaining shallow and small-area contacts. Due to variation in object geometry and scale, models that cannot produce 50 valid patches after resampling are discarded. The final dataset contains 848 objects with 50 viewpoints each and five press depths per valid patch, yielding 212,000 base contact profiles. During training, samples are synthesized on the fly by dynamically applying random wavy masks ($\tilde{M}_\text{wavy}$), correlated color jittering, and random backgrounds ($\tilde{V}_\text{bg}$). This dynamic domain randomization expands the effective dataset and encourages generalizable decoupling rather than memorization of fixed synthetic samples.

\subsection{Reconstruction Framework}

\begin{figure*}[thpb]
  \centering
  \includegraphics[width=0.9\textwidth]{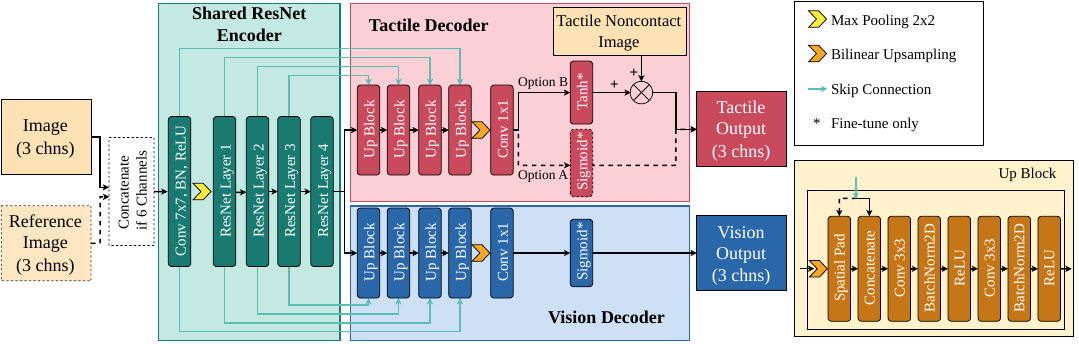}
  \caption{Overview of the dual-stream MuxNet architecture. A shared ResNet-34 encoder processes either a 3-channel fused image or a 6-channel tensor formed by concatenating the multiplexed image with a non-contact reference image. Two task-specific decoders reconstruct the visual and tactile modalities, respectively. The tactile branch supports absolute prediction (Option A) and residual prediction (Option B), where the predicted contact residual is added to the non-contact tactile image. Asterisks ($*$) denote activation functions specific to real-data fine-tuning. Chns: channels. BN: BatchNorm. Conv: Convolution.}
  \label{model}
\end{figure*}

Following the generation of simulated fused data, we propose MuxNet, a dual-stream reconstruction network designed to decouple visual and tactile signals from simulated fused data. To reduce the sim-to-real gap, MuxNet is trained in two stages: large-scale simulation pre-training and real-data fine-tuning with physics-based augmentations. As shown in \cref{model}, MuxNet adopts a modified ResNet34-UNet with a shared encoder and two task-specific decoders.

The network takes a 6-channel input formed by concatenating the current fused image and a non-contact reference image. A ResNet-34 backbone pre-trained on ImageNet is used as the shared encoder. The first convolution is adapted to accept 6 channels while preserving the pre-trained weights. During real-data fine-tuning, the encoder stem (initial convolution, normalization, and pooling) is frozen to stabilize low-level feature extraction and reduce catastrophic forgetting.

The model branches into two symmetric decoder streams with UNet-style upsampling and skip connections. The vision decoder reconstructs the visual image. The tactile decoder is formulated as residual reconstruction. It predicts a contact-induced residual that is added to the non-contact tactile background. Output activations are applied only during real-data fine-tuning: Tanh for tactile residuals, and Sigmoid for vision outputs and absolute tactile predictions. During simulation pre-training, the network optimizes raw logits to avoid early saturation and maintain gradient flow.

Training is governed by a weighted multi-task objective:
\begin{equation}
  \min \text{ } \mathcal{L}_\text{total} = \lambda_\text{t} \mathcal{L}_\text{t} + \lambda_\text{v} \mathcal{L}_\text{v},
  \label{trainObj}
\end{equation}
where $\lambda_\text{t}$ is the weight for the tactile loss $\mathcal{L}_\text{t}$, and $\lambda_\text{v}$ is the weight for the vision loss $\mathcal{L}_\text{v}$.

In the first stage (simulation pre-training), we enforce pixel-accurate supervision:
\begin{align}
\mathcal{L}_\text{v} &= \mathcal{L}_\text{L1}, \label{lVisSim} \\
\mathcal{L}_\text{t} &= \mathcal{L}_\text{L1} + \lambda_\text{grad}\mathcal{L}_\text{grad}, \label{lTacSim}
\end{align}
where $\mathcal{L}_\text{L1}$ denotes the L1 distance between the predictions and ground truth, and $\mathcal{L}_\text{grad}$ computes the L1 distance of the spatial gradients using Sobel-like kernels, forcing the network to sharpen the edges of physical indentations. 

In the second stage (real-world fine-tuning), we introduce physics-based augmentation through random ambient offsets and exposure scaling, and strengthen the loss with structural and perceptual terms:
\begin{align}
\mathcal{L}_\text{v} &= \mathcal{L}_\text{L1} + \lambda_\text{perc}\mathcal{L}_\text{perc}, \label{lVisReal}\\
\mathcal{L}_\text{t} &= \mathcal{L}_\text{L1} + \lambda_\text{grad}\mathcal{L}_\text{grad} + \lambda_\text{SSIM}\mathcal{L}_\text{SSIM} + \lambda_\text{perc}\mathcal{L}_\text{perc}, \label{lTactReal}
\end{align}
where $\mathcal{L}_\text{SSIM}=1-\text{SSIM}$, with SSIM being the Structural Similarity Index Measure \cite{wang2004image}. The perceptual loss $\mathcal{L}_\text{perc}$ follows \cite{johnson2016perceptual} and is computed using feature maps from predefined layers of a frozen VGG-16 network  (relu1\_2, relu2\_2, relu3\_3) \cite{simonyan2014very}. Both stages use AdamW \cite{loshchilov2017decoupled} with a cosine-annealing learning-rate schedule \cite{loshchilov2017sgdr} for optimization. The simulation pre-training stage is trained for 100 epochs, and the real-data fine-tuning stage is trained for 30 epochs.

Model selection is based on a weighted combination of SSIM and Learned Perceptual Image Patch Similarity (LPIPS) \cite{zhang2018unreasonable} across both modalities. The optimal model maximizes the overall score $\mathcal{S}$:
\begin{equation}
\mathcal{S}=\omega_\text{ts}\text{SSIM}_\text{t}-\omega_\text{tl}\text{LPIPS}_\text{t}+\omega_\text{vs}\text{SSIM}_\text{v}-\omega_\text{vl}\text{LPIPS}_\text{v}, \label{score}
\end{equation}
with weights $\omega_\text{ts}=1.0$, $\omega_\text{tl}=0.8$, $\omega_\text{vs}=0.5$, and $\omega_\text{vl}=0.4$. These weights are heuristically fixed before evaluation to prioritize tactile reconstruction for contact-rich manipulation. At each training stage, the score is computed on the corresponding validation split and used only for checkpoint selection.

To isolate the effects of input context and residual learning, we evaluate three variants: SI (Single-Input), DI-AbsT (Dual-Input, Absolute Tactile), and DI-ResT (Dual-Input, Residual Tactile). SI predicts both modalities solely from the current observation. DI-AbsT adds the non-contact reference image as an input and directly predicts the absolute tactile image (Option A in \cref{model}). DI-ResT, our proposed configuration, utilizes the reference image and predicts the visual output and a contact-induced tactile residual added to the non-contact tactile background (Option B). We exclude single-input tactile residual prediction, as the residual is not well-defined and tends to encourage memorization.

\section{Real Data Collection and Model Adaptation}

\begin{figure}[thpb]
  \centering
  \includegraphics[width=0.48\textwidth]{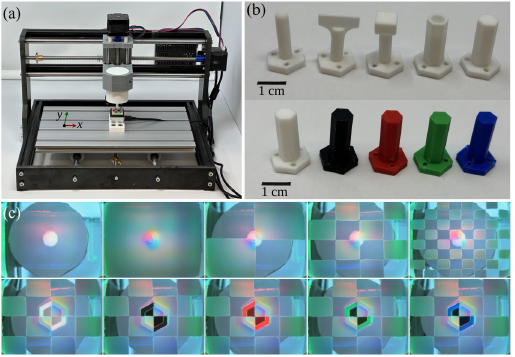}
  \caption{Real-world data collection and dataset overview. (a) Automated 3-axis linear motion platform, where the x,y-axes align with the sensor contact plane for precise spatial sampling. (b) Indenter diversity. Top: five distinct geometries (sphere, edge, square, hollow hexagon, solid hexagon). Bottom: one geometry (hollow hexagon) in five colors (white, black, red, green, blue) to improve visual robustness. Black scale bars correspond to 1 cm. (c) Representative data samples. Top: a white sphere indenter at 1.5 mm depth across five gel configurations. Bottom: a hollow hexagon indenter in five colors captured with the 4$\times$4 patterned gel pad at 1.5 mm depth.}
  \label{fig:calib}
\end{figure}

To support the real-world fine-tuning of MuxNet and reduce the sim-to-real gap, we built an automated tactile data acquisition system, as shown in \cref{fig:calib} (a). The system uses a 3-axis linear motion platform driven by USB-serial G-code to execute repeatable indentations at fixed spatial coordinates.

We collect data with five indenter geometries (sphere, edge, square, solid hexagon, hollow hexagon) to cover smooth point contacts, elongated high-gradient contacts, flat contacts with straight and angled boundaries, and contacts with inner edges, thereby increasing the diversity of both tactile deformation patterns and visual shapes. To promote generalization across visual appearances, each geometry was produced in five colors (white, black, red, green, and blue), resulting in 25 indenters (\cref{fig:calib} (b)). For each indenter, we sample five locations on the sensor surface: the center $(0,0)$ and four peripheral points $(0,-6.4)$, $(0,6.4)$, $(5.2,0)$, and $(-5.2,0)$ mm, using the coordinate system in \cref{fig:calib} (a). At each location, the indenter moves from a non-contact depth of $0$ mm to $1.5$ mm depth in $0.1$ mm steps, capturing continuous gel deformation together with the associated visual occlusion.

We utilized five MuxGel configurations: a fully coated pad (pure tactile), a transparent pad (pure vision), and three multiplexed pads (2$\times$2, 4$\times$4, and 8$\times$8), with representative samples in \cref{fig:calib} (c). Using an identical indentation protocol across pads ensures spatial and depth alignment across all samples. The tactile-only and vision-only pads therefore provide ground-truth targets for supervised fine-tuning of the multiplexed pads. For each multiplexed pad, we capture an initial non-contact reference image under dark conditions as the reference input for the DI-AbsT and DI-ResT variants. Separately, we capture a non-contact tactile background using the fully coated pad, which serves as the additive baseline for residual reconstruction.

This aligned real-world dataset supports the second training stage by adapting the shared encoder and task-specific decoders to sensor-specific effects, including measurement noise, optical scattering, and gel dynamics. We split the dataset into 90\% training and 10\% validation sets and optimize the multi-task objective in \cref{trainObj} with physics-based augmentations. Model selection follows the score $\mathcal{S}$ in \cref{score}, favoring models that preserve both contact-geometry fidelity and external visual reconstruction.

\section{Performance Evaluation on Unseen Objects}

\begin{figure*}[thpb]
  \centering
  \includegraphics[width=0.85\textwidth]{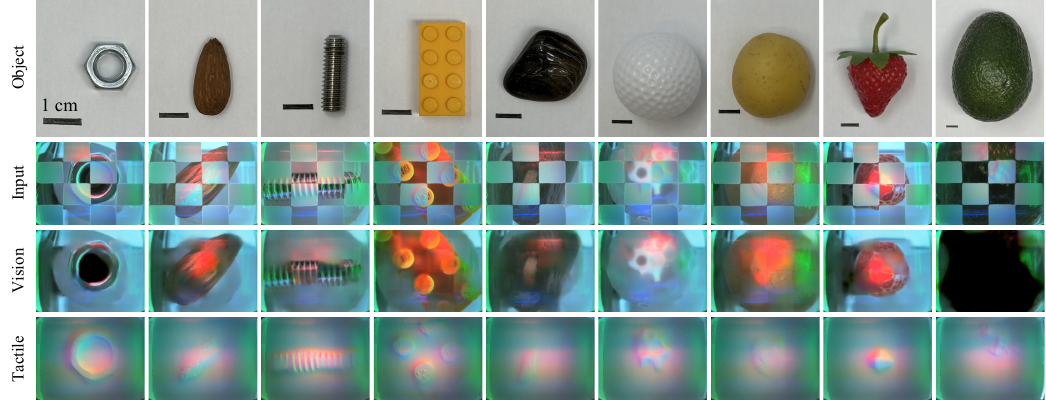}
  \caption{Visuo-tactile reconstruction of real unseen objects using the DI-ResT after real-world fine-tuning. Row 1: Nine objects (screw nut, almond, stub, LEGO block, stone, plastic golf ball, potato, plastic strawberry, avocado) used to test model generalization. Black scale bars correspond to 1 cm. Row 2: Raw multiplexed inputs from the 4$\times$4 checkerboard configuration. Row 3: Reconstructed vision outputs. Row 4: Reconstructed tactile outputs.}
  \label{fig:unseen}
\end{figure*} 

\begin{figure*}[thpb]
  \centering
  \includegraphics[width=0.8\textwidth]{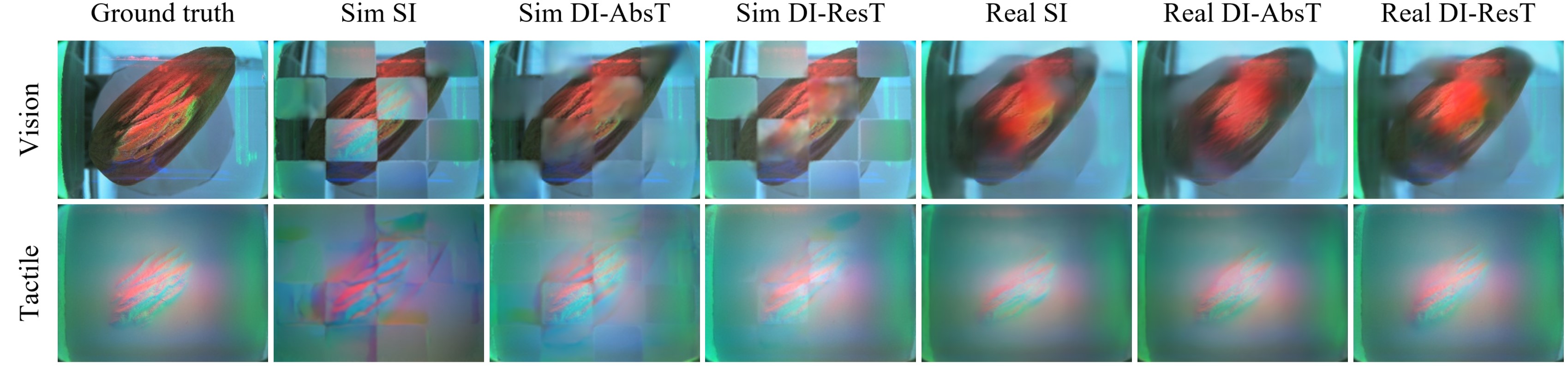}
  \caption{Qualitative reconstruction results on the almond object. Top: vision; bottom: tactile. Left to right: Ground Truth; zero-shot simulation (Sim) models (SI, DI-AbsT, DI-ResT); and real-data fine-tuned models (SI, DI-AbsT, DI-ResT).}
  \label{fig:model}
\end{figure*} 

\begin{table*}[t]
\centering
\caption{Quantitative evaluation of the models on real unseen objects.}
\label{table:model}
\resizebox{\textwidth}{!}{
\begin{tabular}{llcccccccc}
\toprule
\multirow{2}{*}{\textbf{Training Stage}} & \multirow{2}{*}{\textbf{Architecture}} & \multicolumn{4}{c}{\textbf{Tactile Reconstruction}} & \multicolumn{4}{c}{\textbf{Vision Reconstruction}} \\
\cmidrule(lr){3-6} \cmidrule(lr){7-10}
& & \textbf{RMSE $\downarrow$} & \textbf{(1-SSIM) $\downarrow$} & \textbf{LPIPS $\downarrow$} & \textbf{PSNR $\uparrow$} & \textbf{RMSE $\downarrow$} & \textbf{(1-SSIM) $\downarrow$} & \textbf{LPIPS $\downarrow$} & \textbf{PSNR $\uparrow$} \\
\midrule
\multirow{3}{*}{\shortstack[l]{Simulation\\(Zero-shot)}} 
& SI          & 0.0929 & 0.1667 & 0.3749 & 20.6384 & 0.2198 & 0.3651 & 0.3900 & 13.1892 \\
& DI-AbsT     & 0.1260 & 0.1697 & 0.2025 & 18.0412 & 0.1294 & 0.3182 & \textbf{0.3119} & 17.7973 \\
& DI-ResT     & 0.0830 & 0.1227 & 0.1094 & 21.6298 & 0.1553 & 0.3319 & 0.3398 & 16.2726 \\
\midrule
\multirow{3}{*}{\shortstack[l]{Real\\(Fine-tuned)}} 
& SI          & 0.0332 & 0.0956 & 0.0977 & 29.5972 & \textbf{0.1030} & 0.2822 & 0.3557 & \textbf{19.7613} \\
& DI-AbsT     & 0.0770 & \textbf{0.0877} & 0.1082 & 24.6348 & 0.1118 & \textbf{0.2799} & 0.3339 & 19.1292 \\
& \textbf{DI-ResT} & \textbf{0.0287} & 0.0878 & \textbf{0.0489} & \textbf{31.0549} & 0.1058 & 0.3098 & 0.3232 & 19.5489 \\
\bottomrule
\end{tabular}%
}
\\\quad\\
\raggedright
\footnotesize{\textit{Note}: Best results in \textbf{bold}. The vision decoder consistently predicts absolute images. SI: Single-Input. DI: Dual-Input. AbsT: Absolute Tactile. ResT: Residual Tactile. Structural similarity index measure is reported as (1-SSIM).}
\end{table*}

\begin{table*}[t]
\centering
\caption{Quantitative evaluation across MuxGel configurations using the real-data fine-tuned DI-ResT model tested on real unseen objects.}
\label{table:config_ablation}
\resizebox{\textwidth}{!}{%
\begin{tabular}{lcccccccc}
\toprule
\multirow{2}{*}{\textbf{Configuration}} & \multicolumn{4}{c}{\textbf{Tactile Reconstruction}} & \multicolumn{4}{c}{\textbf{Vision Reconstruction}} \\
\cmidrule(lr){2-5} \cmidrule(lr){6-9}
& \textbf{RMSE $\downarrow$} & \textbf{(1-SSIM) $\downarrow$} & \textbf{LPIPS $\downarrow$} & \textbf{PSNR $\uparrow$} & \textbf{RMSE $\downarrow$} & \textbf{(1-SSIM) $\downarrow$} & \textbf{LPIPS $\downarrow$} & \textbf{PSNR $\uparrow$} \\
\midrule
$2 \times 2$ & 0.0301 & 0.0879 & 0.0484 & 30.6372 & 0.1185 & 0.3132 & 0.3457 & 18.5543 \\
$4 \times 4$ & \textbf{0.0285} & \textbf{0.0869} & \textbf{0.0481} & \textbf{31.2080} & 0.1021 & 0.3053 & 0.3136 & 19.8327 \\
$8 \times 8$ & 0.0290 & 0.0891 & 0.0513 & 31.0296 & \textbf{0.0935} & \textbf{0.3015} & \textbf{0.3059} & \textbf{20.5986} \\
\bottomrule
\end{tabular}%
}
\vspace{3pt}
\raggedright
\footnotesize{\quad\\\textit{Note}: Best results among configurations are highlighted in \textbf{bold}. The same DI-ResT model fine-tuned on real training data is evaluated across all configurations. Structural similarity index measure is reported as (1-SSIM).}
\end{table*}

To evaluate the generalization of MuxNet, data is collected on nine unseen objects with diverse colors, textures, and geometric shapes (\cref{fig:unseen}). Each object is mounted on the platform in \cref{fig:calib} (a). Unlike the grid-sampled indenter contacts, each unseen object is placed near the sensor center to avoid premature edge contact caused by uneven object geometry. The platform indents the object from non-contact to 1.5 mm depth in 0.1 mm steps. Data are collected across the same five gel configurations described in \cref{sec:method} A. The tactile-only and vision-only baseline pads serve as the ground-truth tactile and visual targets, respectively, for computing the quantitative results in \cref{table:model}.

We compare SI, DI-AbsT, DI-ResT under zero-shot simulation and real-data fine-tuning training stages. For this evaluation, data from the three multiplexed configurations (2$\times$2, 4$\times$4, and 8$\times$8) are aggregated. Results are shown in \cref{table:model}. RMSE denotes the root mean squared error between reconstructed and ground-truth images, averaged over all pixels, channels, and test samples. SSIM, LPIPS, and PSNR denote Structural Similarity Index Measure, Learned Perceptual Image Patch Similarity, and Peak Signal-to-Noise Ratio, respectively. 

After fine-tuning, DI-ResT achieves the best overall tactile reconstruction performance. In particular, the tactile RMSE of the fine-tuned DI-ResT decreases to 0.0287, improving over the best zero-shot baseline (0.0830). This demonstrates accurate recovery of real contact deformations. The results also show a structural-perceptual trade-off: fine-tuned DI-AbsT and DI-ResT achieve nearly identical tactile (1-SSIM) scores (0.0877 vs. 0.0878), while DI-ResT improves perceptual quality by reducing tactile LPIPS from 0.1082 to 0.0489.

For vision, the zero-shot simulation model (DI-AbsT) attains a lower LPIPS (0.3119) than the fine-tuned models, but with markedly worse PSNR (17.79). This pattern suggests that the simulation-only model produces sharp but structurally inconsistent predictions, likely driven by simulated priors, which can yield favorable perceptual scores while failing to match the real scene. This behavior is also visible in \cref{fig:model}. After real-data fine-tuning, the vision outputs show improved pixel-level agreement with the real targets, as reflected by higher PSNR.

The DI-ResT reconstruction results in \cref{fig:unseen} also reveal several limitations in vision reconstruction. Due to the spatial multiplexing design of MuxGel, the model receives limited color and texture evidence for recovering the hidden visual content, which can lead to over-smoothed predictions. The avocado example represents a typical failure case of near-black artifacts in vision reconstruction, as its large, dark, and weakly textured surface produces visual signals that are close to the low-intensity sensor background. This case shows that the current vision reconstruction is less reliable for low-light and low-texture objects, which we plan to address through lighting-aware augmentation, stronger image priors, and temporal consistency in future work.

Next, we evaluated the reconstruction performance of the fine-tuned DI-ResT model across checkerboard configurations, as detailed in \cref{table:config_ablation}. The results reveal a clear modality trade-off driven by the physical layout of the sensor.

\begin{figure*}[h]
  \centering
  \includegraphics[width=\textwidth]{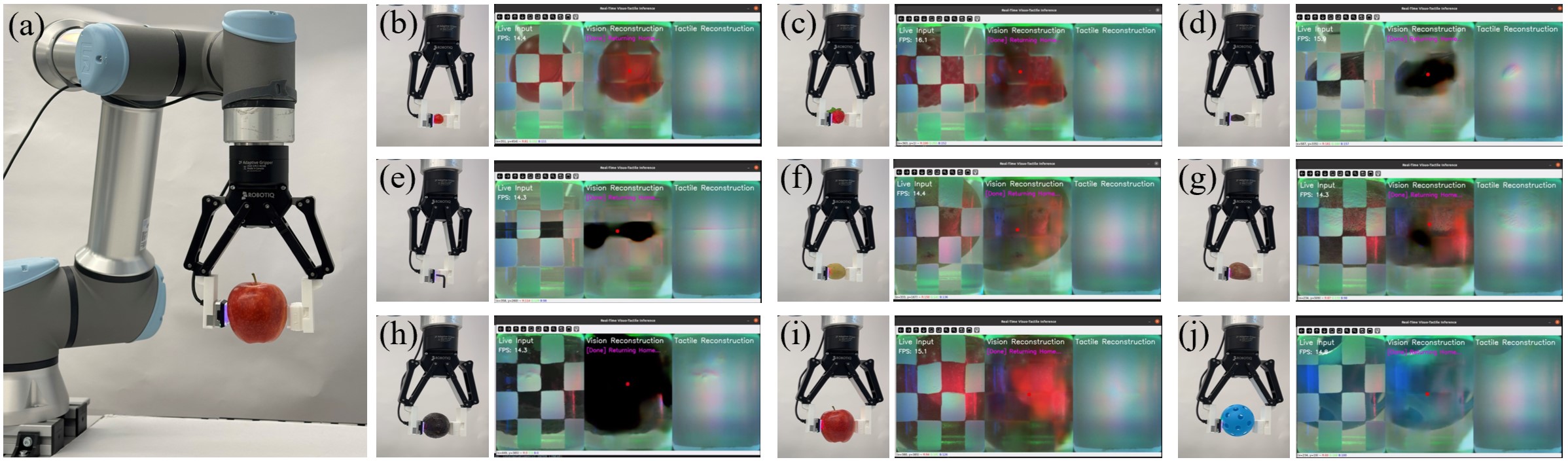}
  \caption{
  Visuo-tactile servoing grasping experiment. (a) Experimental setup featuring the proposed sensor integrated with a Robotiq gripper. (b)-(j) Successful grasp executions across a diverse set of unseen objects. For each object, the left panel depicts the stable physical grasp, while the right panel displays the real-time system interface: capturing the raw sensor input, the reconstructed vision output with contour-center tracking for alignment, and the reconstructed GelSight-style tactile image upon contact. The tested objects include: (b) cherry tomato, (c) plastic strawberry, (d) stone, (e) hex key, (f) potato, (g) red potato, (h) avocado, (i) apple, and (j) pickleball.}
  \label{fig:grasp}
\end{figure*} 

For tactile reconstruction, the $4\times4$ configuration achieves the best performance across metrics, suggesting that its block size provides an effective receptive-field scale for recovering localized contact deformation. Additionally, because the internal illumination of the sensor alters the object's appearance upon contact, the vision stream contains weak contact tactile cues during interaction. The $4\times4$ pattern provides a balanced ratio of coated (tactile) and transparent (vision) regions, avoiding a severe tactile bottleneck while still allowing cross-modal information to be exploited. 

Conversely, for vision reconstruction, the $8\times8$ configuration yields the best performance. As vision depends on global context, smaller and more densely distributed occluded regions minimize the spatial gaps in visual content, which simplifies interpolation for the network.

Tactile sensing prioritizes local high-frequency deformation, while vision prioritizes global structural context, creating a hardware trade-off between the two modalities. Since robotic manipulation relies on localized contact feedback for stability, the $4\times4$ configuration was selected for the subsequent manipulation experiments. The fine-tuned DI-ResT model runs at 17.72 frames per second (FPS), supporting the closed-loop grasping experiment in \cref{sec:visuo-tactile servo}.

\section{Visual-Tactile Servoing Grasping Experiment\label{sec:visuo-tactile servo}}

To evaluate the sensor in a manipulation setting, we designed a dynamic visuo-tactile servoing experiment for robotic grasping. MuxGel was integrated into a Robotiq 2F-140 gripper mounted on a UR16e robotic arm. The robot was constrained to vertical translation (Z-axis) directly above the target objects initially. During the approach phase, the arm descends until the object appears in the sensor's visual field of view. Object localization is achieved via background subtraction and contour extraction. A visual servoing loop continuously computes the object centroid and drives the arm to minimize its vertical offset relative to the sensor center. After alignment, the gripper initiates a gentle closure.

During grasping, the controller uses the reconstructed vision and tactile streams in real time. The vision stream first guides the arm to align the object centroid with the sensor center. Once the centroid falls within a predefined tolerance around the sensor center, the gripper begins closing. Because the Robotiq 2F-140 jaws do not follow a perfectly parallel, purely normal trajectory, closure can induce in-plane object shift. The vision stream continues to track this shift online and adjusts the arm motion to maintain alignment. In parallel, the tactile stream provides contact feedback for closure control. The gripper stops closing when the maximum deformation in the reconstructed tactile depth map exceeds a predefined threshold. This closed-loop strategy aims to establish stable contact while avoiding excessive object compression before lifting. Notably, 3D tactile depth reconstruction (shown in the supplementary video) is performed zero-shot using the standard GelSight Mini repository \cite{gelsight_gsrobotics_2021} without any domain adaptation. This demonstrates that the reconstructed tactile outputs maintain high physical fidelity and are fully compatible with existing tactile processing pipelines. The proposed pipeline was evaluated across nine unseen objects, achieving a 100\% successful grasp rate.

\section{Conclusion}

In this paper, we present MuxGel, a novel spatially multiplexed design that enables simultaneous dual-modal visuo-tactile sensing in vision-based tactile sensors. By using a checkerboard coating with a deep reconstruction model, MuxGel effectively decouples and restores visual and tactile signals from the same camera stream. Experiments show accurate reconstruction and strong generalization to unseen objects. Crucially, MuxGel preserves the pad geometry and mechanical interface of GelSight, enabling plug-and-play integration by replacing only the gel pad, without optical or mechanical changes. Validation within the GelSight platform further confirms its compatibility and ease of deployment.

While this work targets the GelSight format, the spatial multiplexing principle is sensor-agnostic and can be extended to other fully coated vision-based tactile sensors. Future research will focus on optimizing the reconstruction algorithms to handle more complex environmental lighting. Furthermore, we intend to systematically investigate multiplexing designs that vary pattern geometry and the visuo-tactile area ratio to quantify trade-offs in reconstruction and manipulation. We also plan to use the dual-modality signal for downstream tasks such as visuo-tactile pose estimation and closed-loop manipulation in less controlled settings.

% \addtolength{\textheight}{-6cm}   % This command serves to balance the column lengths

%%%%%%%%%%%%%%%%%%%%%%%%%%%%%%%%%%%%%%%%%%%%%%%%%%%%%%%%%%%%%%%%%%%%%%%%%%%%%%%%

%%%%%%%%%%%%%%%%%%%%%%%%%%%%%%%%%%%%%%%%%%%%%%%%%%%%%%%%%%%%%%%%%%%%%%%%%%%%%%%%

%%%%%%%%%%%%%%%%%%%%%%%%%%%%%%%%%%%%%%%%%%%%%%%%%%%%%%%%%%%%%%%%%%%%%%%%%%%%%%%%

% \section*{ACKNOWLEDGMENT}

\bibliographystyle{IEEEtran}
\bibliography{ref}

\end{document}